\documentclass[10pt,twocolumn,letterpaper]{article}

\usepackage{cvpr}              % To produce the CAMERA-READY version
\usepackage{graphicx}
\usepackage{amsmath}
\usepackage{amssymb}
\usepackage{booktabs}
\usepackage{multirow}
\usepackage{xcolor}
\usepackage{array}
\usepackage{xspace}
\usepackage{bbm}

\usepackage[accsupp]{axessibility}

\newcommand{\sh}[1]{\textcolor{black}{#1}}

\newcolumntype{P}[1]{>{\centering\arraybackslash}p{#1}}

\usepackage[pagebackref,breaklinks,colorlinks]{hyperref}

\usepackage[capitalize]{cleveref}
\crefname{section}{Sec.}{Secs.}
\Crefname{section}{Section}{Sections}
\Crefname{table}{Table}{Tables}
\crefname{table}{Tab.}{Tabs.}

\def\cvprPaperID{1500} % *** Enter the CVPR Paper ID here
\def\confName{CVPR}
\def\confYear{2023}

\begin{document}

\title{Relational Context Learning for Human-Object Interaction Detection}

\author{Sanghyun Kim \hspace{0.8cm}  Deunsol Jung \hspace{0.8cm}  Minsu Cho \vspace{1.5mm}\\
Pohang University of Science and Technology (POSTECH), South Korea \\
{\tt\small \{sanghyun.kim, deunsol.jung, mscho\}@postech.ac.kr} \\
\small
\href{http://cvlab.postech.ac.kr/research/MUREN}{\url{http://cvlab.postech.ac.kr/research/MUREN}}
% \hspace{0.8cm}
% \small
% \href{https://github.com/OreoChocolate/MUREN}{\url{https://github.com/OreoChocolate/MUREN}}
}

\maketitle

%%%%%%%%% ABSTRACT
% !TEX root = ../main.tex
\begin{abstract}
% \sh{Human-Object Interaction~(HOI) detection is a challenging task that requires a comprehensive relational understanding to identify the interaction between a human-object pair.}
Recent state-of-the-art methods for HOI detection typically build on transformer architectures with two decoder branches, one for human-object pair detection and the other for interaction classification.
%, has led to significant improvements. 
% Recent state-of-the-art models in Human-Object Interaction~(HOI) detection have shown the effectiveness of the Transformer Encoder-Decoder architecture with two transformer decoders. 
% The disentangled transformer-based HOI methods have the advantage that they can extract the task-specific feature for each sub-task (i.e., human-object pair detection and interaction classification). 
Such disentangled transformers, however, may suffer from insufficient context exchange between the branches and lead to a lack of context information for relational reasoning, which is critical in discovering HOI instances.
In this work, we propose the multiplex relation network (MUREN) that performs rich context exchange between three decoder branches using unary, pairwise, and ternary relations of human, object, and interaction tokens. 
%The multiple relation context information, which contains the unary, pairwise, and triplet relation information, is generated for the context exchange.
% Therefore,
% the mulitpl  contains single, pairwise, and triplet relation context information in an HOI triplet.
%, and propagates the multiple relation context information to the task-specific features.
% The multiple relation context contains information helps the transformer decoders to accommodate rich context information about the HOI triplet with context exchange.
%\sh{We also propose the attentive fusion module which propagates context information to the branches for relational reasoning, selecting the requisite context information for each branch.}
%Therefore, 
The proposed method learns comprehensive relational contexts for discovering HOI instances, % via relational reasoning.
achieving state-of-the-art performance on two standard benchmarks for HOI detection, HICO-DET and V-COCO. 
%and the extensive experiments demonstrate the effectiveness of \ours.

\end{abstract}

%%%%%%%%% BODY TEXT

% !TEX root = ../main.tex
\section{Introduction}
% The goal of Human-Object Interaction (HOI) detection is finding all HOI triplets \textit{$\langle$human,object,interaction$\rangle$} from a given image. Earlier, the conventional methods~\cite{gao2018ican,chao2018learning,qi2018dpnn,gao2020drg,ulutan2020vsgnet,zhang2021spatially,wang2020hetero} solve the HOI detection in two approaches: two-stage and one-stage. 
% In the two-stage methods, they first detect the instances of the human and the object using an off-the-shelf detector and determine the interaction classes for all combinations of human-object pairs.
% In contrast, the one-stage methods directly detect the HOI triplets.
The task of Human-Object Interaction (HOI) detection is to discover the instances of \textit{$\langle$human, object, interaction$\rangle$} from a given image, which reveal semantic structures of human activities in the image.
The results can be useful for a wide range of computer vision problems such as human action recognition~\cite{moon2021integralaction,bretti2021zero,zhang2019structured}, image retrieval~\cite{Wu_2022_CVPR,yoon2021image,gordo2017beyond}, and image captioning~\cite{Wu_2022_caption,yao2018exploring,herdade2019image} where a comprehensive visual understanding of the relationships between humans and objects is required for high-level reasoning.

With the recent success of transformer networks~\cite{vaswani2017attention} in object detection~\cite{detr,zhu2020deformable}, transformer-based HOI detection  methods~\cite{zhang2021cdn,chen2021asnet,kim2022mstr,kim2021hotr,tamura2021qpic,zou2021hoitrans,zhou2022distr}   %which adopt the transformer encoder-decoder architecture~\cite{vaswani2017attention}, 
have been actively developed to become a dominant base architecture for the task. 
Existing transformer-based methods for HOI detection can be roughly divided into two types: single-branch and two-branch.
The single-branch methods~\cite{tamura2021qpic,kim2022mstr,zou2021hoitrans} update a token set through a single transformer decoder and detect HOI instances using the subsequent FFNs directly. 
%detection
% utilize a single transformer decoder a single query set to
As a single transformer decoder is responsible for all sub-tasks (\textit{i.e.,} human detection, object detection, and interaction classification), they are limited in adapting to the different sub-tasks with multi-task learning, simultaneously~\cite{zhang2021cdn}.
%the entangled methods extract the features which contain the context information for detecting the HOI triplet. %each output feature of the
% However, it is challenging to learn the holistic understanding of each sub-task with the multi-task learning, simultaneously~\cite{zhang2021cdn}.
\begin{figure}[t]
\begin{center}
   \includegraphics[]{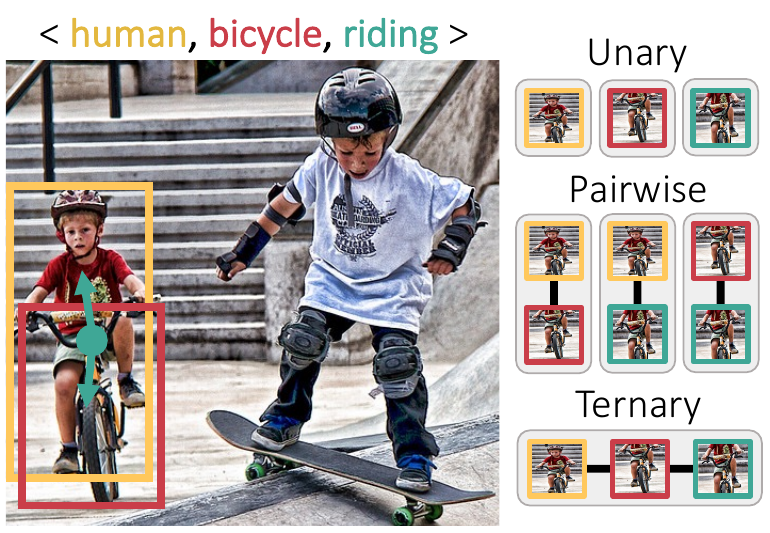}
\end{center}
   \caption{The illustration of relation context information in an HOI instance. We define three types of relation context information in an HOI instance: unary, pairwise, and ternary relation contexts.
   Each relation context provides useful information for detecting an HOI instance.
   For example, in our method, the unary context about an interaction (green) helps to infer that a human (yellow) and an object (red) are associated with the interaction, and vice versa. %(Best viewed in color.)
   Our method utilizes the multiplex relation context consisting of the three relation contexts to perform context exchange for relational reasoning.
   }
   \vspace{-3mm}
\label{fig:teaser}
\end{figure}
%and show significant improvement in an end-to-end manner.
To resolve the issue, the two-branch methods~\cite{zhang2021cdn,kim2021hotr,zhang2022upt,chen2021asnet, zhou2022distr} adopt two separated transformer decoder branches where one detects human-object pairs from a human-object token set while the other classifies interaction classes between  human-object pairs from an interaction token set.
% Since each decoding branch is responsible for different sub-tasks, they are free from above issue.  % using disentangled transformer decoders
However, the insufficient context exchange between the branches prevents the two-branch methods~\cite{kim2021hotr,zhang2021cdn,zhang2022upt} from learning relational contexts, which plays a crucial role in identifying HOI instances.
% However, they ignore the rich context information in the decoding branch due to the disentangled structure. 
% Consequently, they suffer from a lack of context information for the relation reasoning which is a crucial role in identifying the HOI triplet.
% (\textit{e.g.}, the context information of `human' and `surfboard' help to infer `surf' interaction).
Although some methods~\cite{chen2021asnet, zhou2022distr} tackle this issue with additional context exchange, they are limited to propagating human-object context to interaction context.
To address the problem, we introduce the  \textbf{MU}tiplex \textbf{RE}lation \textbf{N}etwork~(MUREN) that performs rich context exchange using unary, pairwise, and ternary relations of human, object, and interaction tokens for relational reasoning.
%for context exchange between each decoding branch.
As illustrated in Figure~\ref{fig:teaser}, we define three types of relation context information in an HOI instance: unary, pairwise, and ternary, each of which provides useful information to discover HOI instances.
% These three relation context information helps to derive a comprehensive visual understanding of the HOI triplet.
The ternary relation context gives holistic information about the HOI instance while the unary and pairwise relation contexts provide more fine-grained information about the HOI instance. For example, as shown in Figure~\ref{fig:teaser}, the unary context about an interaction (\textit{e.g.,} `riding') helps to infer which pair of a human and an object is associated with the interaction in a given image, %and vice versa. % \sh{since it contains the information for associating the
%human and object.}
and the pairwise context between a human and an interaction (\textit{e.g.,} `human' and `riding') helps to detect an object (\textit{e.g.,} `bicycle').
% In the proposed method, we generate the multiplex relation context that contains the three relation context information \sh{for context change between the branches.} 
Motivated by this, our multiplex relation embedding module constructs the context information that consists of the three relation contexts, thus effectively exploiting their benefits for relational reasoning.
%to derive relation reasoning about the HOI triplet.
% To utilize these three relation context information, we generate the multiple relation context information which contains, the unary, the pairwise and the triplet relation in the HOI triplet. %to enrich the task-specific features.
Since each sub-task requires different context information for relational reasoning, our attentive fusion module selects requisite context information for each sub-task from multiplex relation context and propagates the selected context information for context exchange between the branches.
Unlike previous methods~\cite{kim2021hotr,chen2021asnet,zhang2021cdn,zhou2022distr}, we adopt three decoder branches which are responsible for human detection, object detection, and interaction classification, respectively.
Therefore, the proposed method learns discriminative representation for each sub-task.

%multiplex relation contexts 
%\sh{Therefore, the proposed method learns comprehensive relational contexts for discovering the HOI instances via context exchange among three branches using the multiplex context information and the attentive fusion module.}
We evaluate~MUREN~on two public benchmarks, HICO-DET~\cite{hico} and V-COCO~\cite{vcoco}, showing that MUREN~achieves state-of-the-art performance on two benchmarks.
The ablation study demonstrates the effectiveness of the multiplex relation embedding module and the attentive fusion module.
% We use three transformer decoders, which extract task-specific feature for each sub-task, rather than single decoder to leverage the benefits of disentangled feature. 
% we utilize the single, pair-wise and triplet information (i.e., multiplex relation embedding), simultaneously, to contextualize a HOI triplet.
% The triplet information gives holistic context information about a HOI triplet.
% The single and pair-wise information give more fine-grained context information about each sub-task.
% For example, a pair of human and interaction context (e.g., 'human' and 'hitting') helps to detect a object (e.g.,'ball'). 
% A single context information of interaction (e.g., 'riding') gives a good prior to perceive which a human and a object in a given image is associated with the interaction.
% The attentive fusion module is used to perform the message passing between each sub-task using the multiplex relation embedding.
% In the module, the multiplex relation embedding is refined into task-specific message guided with the disentangled features and channel attention is used to allows the multiplex relation embedding to capture the context for the sub-task.
Our contribution can be summarized as follows:
\begin{itemize}
%   \item We propose a novel HOI detection method, dubbed MUREN, that performs effective context exchange for relational reasoning.
%   a three-branch disentangled transformer-based architecture to leverage the multiple relation context information. %the benefits of disentangled transformer-based HOI method.
  \item We propose multiplex relation embedding module for HOI detection, which generates context information using unary, pairwise, and ternary relations in an HOI instance.
  \item We propose the attentive fusion module that effectively propagates requisite context information for context exchange.
  \item We design a three-branch architecture to learn more discriminative features for sub-tasks, \textit{i.e.}, human detection, object detection, and interaction classification.
  \item Our proposed method, dubbed MUREN, outperforms state-of-the-art methods on HICO-DET and V-COCO benchmarks.
  
\end{itemize}
% !TEX root = ../main.tex
\section{Related Work}
\subsection{CNN-based HOI Methods.}
Previous CNN-based HOI methods can be categorized into two groups: two-stage methods and one-stage methods.
Two-stage HOI methods~\cite{li2019tin,gao2020drg,gao2018ican,li2020IDN,qi2018dpnn,ulutan2020vsgnet,wang2020hetero,zhang2021spatially,hou2020vcl} first detect the human and the object instances using an off-the-shelf detector~(\textit{e.g.,} Faster R-CNN~\cite{ren2015faster}) and predict the interaction between all possible pairs of a human and an object. 
To create discriminative instance features for HOI detection, they additionally utilize spatial features~\cite{gao2018ican,spaital_learning,li2019tin}, linguistic features~\cite{liu2020consnet,gao2020drg}, and human pose features~\cite{li2019tin,gupta2019no} with visual features.
Some approaches~\cite{qi2018dpnn,gao2020drg,wang2020hetero,ulutan2020vsgnet,zhang2021spatially} utilize the graph structure and exchange the context information of the instance features for relational reasoning between the nodes.
DRG~\cite{gao2020drg} proposes human-centric and object-centric graphs to perform context exchange focused on relevant context information. SCG~\cite{zhang2021spatially} transforms and propagates the context information to the nodes in a graph conditioned on spatial relation.
% We also exploit the contextual information to make discriminative features.
% Different from previous work, we utilize the multiple relation context information in an HOI triplet for the relation reasoning.
On the other hand, previous one-stage HOI methods~\cite{kim2020uniondet,liao2020ppdm,fang2021dirv} detect human-object pairs and classify the interactions between human-object pairs in an end-to-end manner.
These methods utilize the interaction region to match the interaction and a pair of a human box and an object box. 
UnionDet~\cite{kim2020uniondet} proposes a union-level detector to find the union box of human and object for matching a human-object pair.
PPDM~\cite{liao2020ppdm} detects interaction centers and points to the center point of the human and object box to predict HOI instances.

%typically depend on predefined
%interaction areas for interaction prediction.

% \noindent
% \textbf{One-stage HOI methods.} Previous one-stage 

% \hspace{-1cm}
\begin{figure*}[t]
% \begin{center}
   \includegraphics[width=0.94\textwidth]{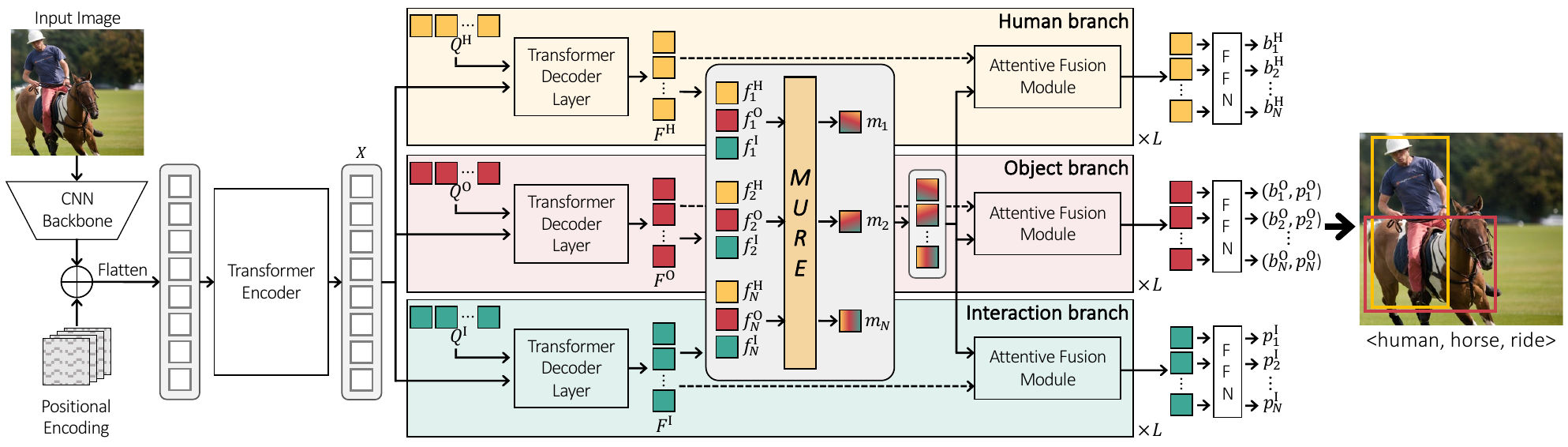}
% \end{center}
   \caption{The overall architecture of MUREN. The proposed method adopts three-branch architecture: human branch, object branch, and interaction branch. Each branch is responsible for human detection, object detection, interaction classification. The input image is fed into the CNN backbone followed by the transformer encoder to extract the image tokens. A transformer decoder layer in each branch layer extracts the task-specific tokens for predicting the sub-task. The MURE takes the task-specific tokens as input and generates the multiplex relation context for relational reasoning. The attentive fusion module propagates the multiplex relation context to each sub-task for context exchange. The outputs at the last layer of each branch are fed into to predict the HOI instances.}
   \vspace{-3mm}
\label{fig:framework}
\end{figure*}

\subsection{Transformer-based HOI Methods.} 
% Recently, the transformer~\cite{vaswani2017attention} has been successfully adapted and brings significant improvements in object detection~\cite{detr,zhu2020deformable}. Inspired by this, 
Inspired by DETR~\cite{detr}, a number of work~\cite{zhou2022distr,zou2021hoitrans,chen2021asnet,tamura2021qpic,kim2022mstr,zhang2022upt,kim2021hotr} have adopted the transformer-based object detector to solve HOI detection. 
They can be divided into two folds: single-branch and two-branch methods.
The single-branch methods~\cite{tamura2021qpic,kim2022mstr,zou2021hoitrans} predict the HOI instances with a single transformer decoder.
MSTR~\cite{tamura2021qpic} utilizes multi-scale features to extract discriminative features for the HOI instances.
In contrast, two-branch methods~\cite{kim2021hotr,chen2021asnet,zhou2022distr,zhang2022upt,zhang2021cdn} adopt two transformer decoder branches, one is responsible for human-object pair detection and the other for interaction classification.
HOTR~\cite{kim2021hotr} detects the instances in an image in detection branch and predicts the interaction with additional offsets to associate humans and objects in interaction branch.
Although they extract discriminative features for each sub-task, there is no context exchange for relational reasoning, bringing performance degradation in HOI detection.
To alleviate this, AS-NET~\cite{chen2021asnet} and DisTR~\cite{zhou2022distr} perform the message passing for relational reasoning between two branches. 
However, they only propagate human-object context information for interaction classification.
% !TEX root = ../main.tex
% HOTR~\cite{kim2021hotr}, which is one of the disentangled methods, use two transformer decoders which extract the disentangled feature where one is responsible for detecting the instance and another is interaction classification.
% Although they can focus on each task due to the disentangled features, there is no relation reasoning which causes performance degradation in HOI detection. 
% AS-NET~\cite{chen2021asnet} and DisTR~\cite{zhou2022distr} are perform the message passing to perform relation reasoning with disentangled structure.
% However, they only consider one-way message passing where the instance context is propagated to the feature for interaction classification.
% PST~\cite{dong2021pst} propose the part and sum part query to represent the features for the sub-tasks and an HOI instance, respectively, performing the context exchange between the sum query and the part queries.
% However, they only consider one type of relation context information when propagating the context information.
In this paper, we exchange the context among branches with the multiplex relation context.
The multiplex relation context, which considers all relation contexts in an HOI instance, gives relational semantics for relational reasoning.
We also extract more discriminative features for each sub-task via three-branch.

% levering the advantage of the disentangled feature. 
% entangled

%disentangled
%two-stage

% !TEX root = ../main.tex
% \vspace{-2}
\section{Problem Definition}
% HOI detection aims to predict a set of HOI instances. Each instance is composed of a bounding box of human $\mathbf{b}^H_{i}\in \mathbb{R}^{4}$, a bounding box of object $\mathbf{b}^O_{i}\in \mathbb{R}^{4}$, class probability of object $\mathbf{p}^{O}_{i}\in \mathbb{R}^{|\mathcal{O}|}$, and class probability of interaction $\mathbf{p}^I_{i}\in \mathbb{R}^{|\mathcal{I}|}$, where $|\mathcal{O}|$ and $|\mathcal{I}|$ indicate the number of object classes and the number of interaction classes, respectively.

Given an input image, the goal of HOI detection is to predict a visually-grounded set of HOI instances for object classes $\mathcal{O}$ and interaction classes $\mathcal{I}$. %an image with graph structure, 
%the detected objects correspond to the nodes in the graph while pairwise relations between them describe the edges. 
An HOI instance consists of four components: a bounding box of human $\mathbf{b}^\mathrm{H}_{i}\in \mathbb{R}^{4}$, a bounding box of object $\mathbf{b}^\mathrm{O}_{i}\in \mathbb{R}^{4}$, a one-hot vector of object label $\mathbf{c}^{\mathrm{O}}_{i}\in \{ 0, 1\}^{|\mathcal{O}|}$, and a one-hot
vector of interaction label $\mathbf{c}^\mathrm{I}_{i}\in \{ 0, 1\}^{|\mathcal{I}|}$, where $|\cdot|$ denotes the size of a set. % the number of object classes and the number of interaction classes, respectively. 
 The output of HOI detection is thus expressed by a set of HOI instances \{$(\mathbf{b}^\mathrm{H}_{i}, \mathbf{b}^{\mathrm{O}}_{i}, \mathbf{c}^{\mathrm{O}}_{i}, \mathbf{c}^\mathrm{I}_{i})\}$. 

% $\mathbf{p}^{O}_{i} \in \{ 0, 1\}^{|\mathcal{O}|}$

% $G=(\mathcal{B},\mathcal{O},\mathcal{R})$.   
% $\mathcal{B}=\{b_1, \dots, b_n\}$ is a set of object bounding boxes, where $b_i \in [0,1]^4$; 
% $\mathcal{O}=\{o_1, \dots, o_n\}$ is a set of labels assigned to each $b_i$, where $o_i \in \mathcal{C}$ and $\mathcal{C}$ is a set of predefined object classes;  
% $\mathcal{R} = \{r_1, \dots, r_k\}$ is a set of relationships. 
% Each relationship $r_k$ is a triplet of a subject $(b_i, o_i)$, an object $(b_j, o_j)$ and a predicate label $p_{ij} \in \mathcal{P}$, which represents the relationship between the subject $i$ and the object $j$, 
% where $\mathcal{P}$ is a set of predefined predicate classes.

\section{Method}

% HOI detection aim to predict a set of HOI instances \{$(\mathbf{b}^H_{i}, \mathbf{b}^{O}_{i}, \mathbf{p}^{O}_{i}, \mathbf{p}^I_{i})\}_{i=1}^{N}$ where $\mathbf{b}^{H}_{i},\mathbf{b}^{O}_{i} \in \mathbb{R}^{4}$, $\mathbf{p}_{i}^{O}\in \mathbb{R}^{|\mathcal{O}|}$, $\mathbf{p}_{i}^{I} \in \mathbb{R}^{|\mathcal{I}|}$ are a bounding box of human, a bounding box of object, class probability of object, and class probability of interaction. $|\mathcal{O}|$ and $|\mathcal{I}|$ indicate the number of object classes and the number of interaction classes, respectively.

% HOI detection aims to predict a set of HOI instances. Each instance is composed of a bounding box of human $\mathbf{b}^H_{i}\in \mathbb{R}^{4}$, a bounding box of object $\mathbf{b}^O_{i}\in \mathbb{R}^{4}$, class probability of object $\mathbf{p}^{O}_{i}\in \mathbb{R}^{|\mathcal{O}|}$, and class probability of interaction $\mathbf{p}^I_{i}\in \mathbb{R}^{|\mathcal{I}|}$, where $|\mathcal{O}|$ and $|\mathcal{I}|$ indicate the number of object classes and the number of interaction classes, respectively.

The proposed network, MUREN, is illustrated in Figure~\ref{fig:framework}.
Given an input image, it extracts image tokens via a CNN backbone followed by a transformer encoder. 
The image tokens are fed to three independent branches to perform three sub-task: human detection, object detection, and interaction classification.
In each branch, a transformer decoder layer refines $N$ learnable tokens using the image tokens as keys and values to extract task-specific tokens.
Using the task-specific tokens of each branch, our multiplex relation embedding module~(MURE) generates the context information for relational reasoning. The attentive fusion module then integrates the context information across the task-specific tokens for human, object, and interaction branches, propagating the results to the next layer.
After repeating this process for $L$ times, FFNs predict the set of HOI instances.
In the remainder of this section, we explain the details of each component in MUREN.

\subsection{Image Encoding}
% \vspace{-1.5mm}
Following the previous work~\cite{detr,tamura2021qpic,zou2021end}, we use a transformer encoder with a CNN backbone to extract image tokens.
The CNN backbone takes an input image to extract an image feature map.
% Given an image, an image feature map is generated by using the CNN backbone.
The image feature map is fed into $1\times1$ convolution layer to reduce the channel dimension to $D$, and the positional encoding~\cite{detr} is added to the image feature map to reflect the spatial configuration of the feature map. 
The feature map is then tokenized by flattening and fed into  
%to form a set of image tokens.
the transformer encoder 
%takes \sh{the flattened image feature map} as inputs 
to produce image tokens $\mathbf{X} \in \mathbb{R}^{T \times D}$ for the subsequent networks, where $T$ and $D$ are the number of the image tokens and the channel dimension, respectively.

% \vspace{-1mm}
\subsection{HOI Token Decoding}
% \vspace{-1mm}
% We adopt the disentangled transformer decoders to extract the task-specific feature for each sub-task.
% We utilize three separated decoders to contextualize the feature representations for each sub-task.
Different from previous two-branch methods~\cite{kim2021hotr,zhou2022distr,chen2021asnet}, we design an architecture consisting of three branches which is responsible for human detection, object detection, and interaction classification, respectively.
% The $B^H,B^O$ and $B^I$ provide discriminative features for predicting human detection, object detection/classifiaction and interaction classification.
% The $B^H,B^O$ and $B^I$ are extract the task-specific feature for the human detection, object detection and interaction classification, respectively.
Each branch $\tau$, consisting of $L$ layers, takes the learnable tokens $\mathbf{Q}^{\tau}=\{\mathbf{q}^{\tau}_{i}\}^{N}_{i=1}$ and the image tokens $\mathbf{X}$ as inputs , where $\tau \in \{\mathrm{H, O, I}\}$ indicates human, object, and interaction respectively. At each layer, $\mathbf{Q}^{\tau}$ is refined through a transformer decoder layer followed by a MURE module and an attentive fusion module. Specifically, the three branches take learnable tokens $\mathbf{Q}^\mathrm{H},\mathbf{Q}^\mathrm{O},\mathbf{Q}^\mathrm{I}\in\mathbb{R}^{N\times D}$ for human, object, and interaction branches, respectively.
% respective human $\mathbf{Q}^H$, object $\mathbf{Q}^O$, interaction $\mathbf{Q}^I$, where $\mathbf{Q}^H,\mathbf{Q}^O,\mathbf{Q}^I\in\mathbb{R}^{N\times D}$ and $N$ is the number of queries.
% The $\mathbf{Q}^H,\mathbf{Q}^O$, and $\mathbf{Q}^I$ are refined in each branch and utilized to predict each sub-task.
% To refine the token set, $l$-th layer of each branch first generates the task-specific \sh{token set} $\mathbf{F}^\tau_{(l)}$ which contains the context information for predicting each sub-task using a transformer decoder layer as follows:
In $l$-th layer of the branch $\tau$, a transformer decoder layer $\text{Dec}^\tau_{(l)}$ updates $\mathbf{Q}^\tau _{_{(l-1)}}$, the output of previous layer of the branch $\tau$, by attending $\mathbf{X}$ to generate task-specific tokens $\mathbf{F}^\tau_{(l)}=\{\mathbf{f}^{\tau}_{(l),i}\}^{N}_{i=1}$ which contain the context information for predicting a sub-task which the branch $\tau$ is responsible for:

\begin{equation}
    \mathbf{F}^\tau_{(l)} = \text{Dec}^\tau_{(l)}(\mathbf{Q}^\tau _{_{(l-1)}}, \mathbf{X}),
    \label{eq:task-specific}
\end{equation}
where $\text{Dec}(q,kv)$ denotes a transformer decoder layer.

\subsection{Relational Contextualization}
% As mentioned above, the context information of each branch is crucial for relational reasoning to identify HOI instances. 

\begin{figure}[t!]
\centering
   \includegraphics[width=0.45\textwidth]{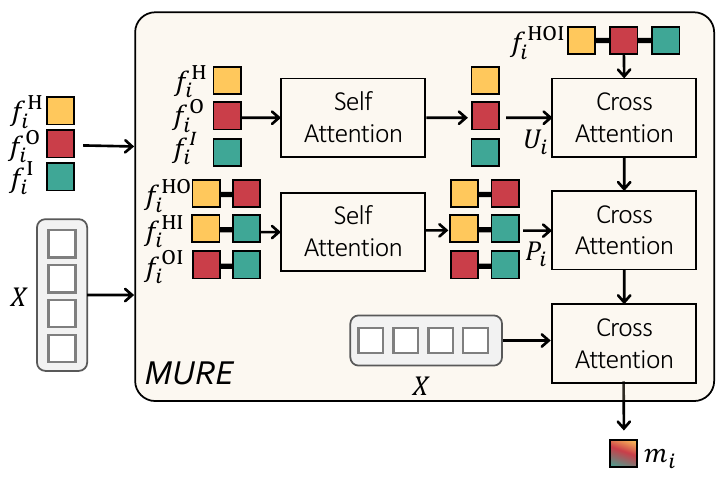}
   \caption{The architecture of the multiplex relation embedding module~(MURE). MURE takes $i$-th task-specific tokens and the image tokens as input, and embed the unary and pairwise relation contexts into the ternary relation context. The multiplex relation context, the output of MURE, is fed into subsequent attentive fusion module for context exchange.}
\label{fig:mure_fig}
\vspace{-0.4cm}
\end{figure}

%across different branch outputs for effective relational reasoning.
% Our transformer decoder layer in each decoding branch generates the task-specific features for predicting each sub-task.
% However, these features cannot be aware of the context information about each other, which is important to identifying the HOI triplets via relation reasoning.
As mentioned above, relational reasoning is crucial to identify HOI instances.
However, since the task-specific tokens are generated from the separated branches, the tokens suffer from a lack of relational context information.
To mitigate this issue, we propose multiplex relation embedding module~(MURE) which generates multiplex relation context for relational reasoning.
The multiplex relation context contains the unary, pairwise, and ternary relation contexts to exploit useful information in each relation context, as shown in Figure~\ref{fig:mure_fig}.
% As shown in Figure~\ref{fig:mure_fig}, the multiplex relation context information is generated by considering the unary, the pairwise, and the ternary relations in an HOI instance.
% The multiple relation context information contains the unary, the pairwise and the triplet relation context information in an HOI triplet.
% Therefore, the task-specific features deduce the comprehensive visual understanding for the HOI triplet via the relation reasoning. 
%to communicate the contextual information to the disentangled features.

% Specifically, the MURE first constructs the ternary relation context $\mathbf{f}^{HOI}_{i}$ for $i$-th HOI instance \sh{by concatenating $i$-th task-specific tokens of each branch $\mathbf{f}^\tau_i$ }:
% \sh{by concatenating $\mathbf{f}^\tau_i$}, which is $i$-th element in the task-specific feature $\mathbf{F}^{\tau}$:
% The MURE takes the task-specific features as input and first generates the triplet relation context information: 
Specifically, the MURE first constructs the ternary relation context $\mathbf{f}^{\mathrm{HOI}}_{i}\in \mathbb{R}^{D}$ for $i$-th HOI instance by concatenating each $\mathbf{f}^\tau_i$ followed by an MLP layer.
\begin{equation}
     \mathbf{f}^{\mathrm{HOI}}_{i} = \text{MLP}([\mathbf{f}^\mathrm{H}_i;\mathbf{f}^\mathrm{O}_i;\mathbf{f}^\mathrm{I}_i]), \label{eq:triplet}
\end{equation}
where $[\cdot;\cdot]$ is a concatenation operation.
We omit the subscript $l$ for the sake of simplicity.
% $\mathbf{f}^\tau_i$ is $i$-th element in the task-specific \sh{token set} $\mathbf{F}^{\tau}$. 
Since the ternary relation takes the overall understanding of each sub-task into account, it gives holistic context information about the HOI instance.
% The triplet information takes into account the overall understanding of each sub-task 
% Since the triplet information contains holistic contextual information about the HOI triplet, 
% they  gives coarse to task-specific feature
On the other hand, since the unary and the pairwise relations take a fine-grained level understanding of each sub-task into account, they give the fine-grained context information about the HOI instance.
To exploit both holistic and fine-grained context information, we embed the unary and the pairwise relation contexts within the ternary relation context with a sequential manner.
% Specifically, we utilize a self-attention~(SA) mechanism to generate the unary and the pairwise relation context information from the task-specific features.
% Then, the unary relation context information is embedded in the triplet relation context information using a  cross-attention~(CA) mechanism.
% and a cross-attention~(CA) mechanism to refine the context information of the triplet relation.
% In detail, A set of unary relation context information $U_{i}$ for $i$-th HOI triplet is generated a self-attention on a set of $i$-th task-specific features $\{f^H_i,f^O_i,f^I_i\}$ as Eq.~\ref{eq:sa_single}.

% to generate a set of unary relation context information $U_{i}$ for $i$-th HOI triplet,
In detail, we apply a self-attention on a set of $i$-th task-specific tokens $\{\mathbf{f}^\mathrm{H}_i,\mathbf{f}^\mathrm{O}_i,\mathbf{f}^\mathrm{I}_i\}$ to consider the unary relation for $i$-th HOI instance as Eq.~\ref{eq:sa_single}.
%for $i$-th HOI instance $U_i$ as Eq.~\ref{eq:sa_single}.
Then, the unary-relation context $U_i$ is embedded into ternary relation context using a cross-attention as Eq.~\ref{eq:ca_single}:
%  to enrich the triplet relation context information with the unary relation context information
% Specifically, The self-attention is applied to a set of $i$-th HOI unary information $Q^{S}_{i}=\{q^H_i,q^O_i,q^I_i\}$ to encode the single information in the HOI triplet.
% Then, the single information is encoded into triplet information using the cross-attention as follows:
\begin{align}
    U_i &= \text{SelfAttn}(\{\mathbf{f}^\mathrm{H}_i,\mathbf{f}^\mathrm{O}_i,\mathbf{f}^\mathrm{I}_i\}), \label{eq:sa_single} \\
   \tilde{\mathbf{f}}^{\mathrm{HOI}}_{i} &= \text{CrossAttn}(\mathbf{f}^{\mathrm{HOI}}_i,
   U_i),
   \label{eq:ca_single}
\end{align}
% Let
% \begin{gather}
%     \hat{\mathbf{F}} = \text{concat}\left({\mathbf{f}^H_i, \mathbf{f}^O_i, \mathbf{f}^I_i} \right) \in \mathbb{R}^{3 \times D} \\ 
%     \text{SA}^{(h)}(\hat{\mathbf{F}})_{\tau} = \text{softmax}(\mathbf{A}^{(h)})_{\tau} \hat{\mathbf{F}}\mathbf{W}^{(h)}_{\text{v}} \\
%     \mathbf{A}^{(h)}_{\tau} = \hat{\mathbf{F}}\mathbf{W}^{(h)}_{\text{q}} \cdot (\hat{\mathbf{F}}\mathbf{W}^{(h)}_{\text{k}})^{\top}
% \end{gather}
where we denote $\text{SelfAttn}(\cdot)$ as a self-attention operation and $\text{CrossAttn}(q,kv)$ as a cross-attention operation for simplicity.
% we utilize a self-attention~(SA) mechanism and a cross-attention (CA) mechanism.
%For the simplicity, we denote the $\text{SA}(x)=\text{FFN}(\MHA())$.
To embed the pairwise relation context within the ternary relation context, we extract the pairwise features of $\mathbf{f}^{\mathrm{HO}},\mathbf{f}^{\mathrm{HI}},\mathbf{f}^{\mathrm{OI}}\in \mathbb{R}^{D}$ for respective human-object, human-interaction, object-interaction relation as follows:
\begin{align}
    \mathbf{f}^{\mathrm{HO}}_i &= \text{MLP}([\mathbf{f}^\mathrm{H}_i;\mathbf{f}^\mathrm{O}_i]), \\
    \mathbf{f}^{\mathrm{HI}}_i &= \text{MLP}([\mathbf{f}^\mathrm{H}_i;\mathbf{f}^\mathrm{I}_i]), \\
    \mathbf{f}^{\mathrm{OI}}_i &= \text{MLP}([\mathbf{f}^\mathrm{O}_i;\mathbf{f}^\mathrm{I}_i]).
    \label{eq:pariwise}
\end{align}
%where $ab \in \{HO,HI,OI\}$.
% The same as the unary relation context information, we a set of pairwise relation context information for $i$-th HOI instance, is generated using a self-attention and embedded into the triplet relation context information using a cross-attention as follows:
Similar to the above,
we apply the self attention on a set of pairwise features to consider the pairwise relation for $i$-th HOI instance, and the cross attention to embed the pairwise relation contexts within ternary relation context:
\begin{align}
    P_i &= \text{SelfAttn}(\{\mathbf{f}^{\mathrm{HO}}_i,\mathbf{f}^{\mathrm{HI}}_i,\mathbf{f}^{\mathrm{OI}}_i\}), \label{eq:sa_pair} \\ 
    \hat{\mathbf{f}}^{\mathrm{HOI}}_{i} &= \text{CrossAttn}(\tilde{\mathbf{f}}^{\mathrm{HOI}}_{i},P_i). \label{eq:ca_pair}
\end{align}

% To encode the pairwise information, we first generate all possible pairwise feature with the Multi-Layer Perceptrons~(MLP):  
% \begin{equation}
% \begin{split}
%     f^{HO}_i = \text{MLP}([q^H_i;q^O_i]), \\ 
%     f^{HI}_i = \text{MLP}([q^H_i;q^O_i]), \\
%     f^{OI}_i = \text{MLP}([q^O_i;q^I_i]).
% \end{split}
% \end{equation}

Finally, the $\hat{\mathbf{f}}^{\mathrm{HOI}}_{i}$ is transformed to generate the multiplex relation context $\mathbf{m}_i$ as follows by attending the image tokens $\mathbf{X}$:
% Finally, a embedding feature to communicate between the disentangled features is generated:
\begin{equation}
    \mathbf{m}_i  = \text{CrossAttn}(\hat{\mathbf{f}}^{\mathrm{HOI}}_{i},\mathbf{X}).
    \label{eq:mure_image_feat}
\end{equation}

\sh{It is noteworthy that our high-order (ternary and pairwise) feature functions have a form of non-linear function, \textit{i.e.}, MLP, on top of a tuple of multiple inputs, which is not reducible to a sum of multiple functions of individual lower-order inputs in general. Such a high-order feature function thus can learn the structural relations of the inputs in the tuple, considering all the inputs jointly. For example, a ternary function of three coordinates $f(a, b, c)$ can compute the angle feature between {$\overline{ab}$} and {$\overline{ac}$}, which cannot be computed by an individual unary function, $g(a)$, $g(b)$, or $g(c)$ as well as their linear combination. In a similar vein, our ternary feature functions, \textit{i.e.}, Eq.~\ref{eq:triplet}, can effectively learn to capture structural relations which are not easily composable from unary and pairwise feature functions.
}
%of (H, O, I) feature triplets, which are not easily composable from unary and pairwise feature functions.}

% Since the multiple relation context information contains the unary, the pairwise and the triplet relation context information in an HOI triplet, the task-specific features deduce the comprehensive visual understanding for the HOI triplet via the relation reasoning. 

\subsection{Attentive Fusion}
Our attentive fusion module aims to propagate the multiplex relation context to the task-specific tokens for context exchange.
%To deduce the relation reasoning via comprehensive visual understanding for detecting an HOI triplet, it is important to recognize the context information about the HOI triplet.
% \sh{We propose the attentive fusion module to propagate generated multiple relation context information in MURE.}
% The multiple relation context information is propagated to the task-specific features for relation reasoning in the attentive fusion module.
% Specifically, the proposed attentive fusion module selects requisite context information for each sub-task from the multiple relation context information using the channel attention.
% Then, the selected context information is fused with the task-specific features to generate $Q^a_{(l)}$.
Since each sub-task requires different context information for relational reasoning, the multiplex relation context is transformed using MLP with each task-specific token to propagate the context information conditioned on each sub-task.
% \sh{Specifically, the proposed attentive fusion module propagates the multiple relation context information conditioned on each task-specific feature, since each task requires different context information to perform relation reasoning.}
We further utilize the channel attention to select the requisite context information for each sub-task.
Then, the refined tokens $\mathbf{Q}^{\tau}_{(l)}$, the output of $l$-th layer of branch~$\tau$, is generated by propagating the requisite context information to the task-specific tokens $\mathbf{F}^\tau_{(l)}$.
% \sh{Formally, a task-specific feature $f^{a}_{i}$ and the multiple relation context information $m_{i}$ are concatenated and fed into MLP layer to}
%Specifically, the multiple relation context information is transformed conditioned on each sub-task, and the channel attention is applied to select the requisite context information for each sub-task.
% Formally, a task-specific feature $f^{a}_{i}$ and the multiple relation context information $m_{i}$ are concatenated and fed into MLP layer to transform multiple relation context information \sh{to select the context information conditioned on the task.}
% % \sh{A channel attention $\alpha$ is also utilized to select the context information.}
% % and compute a channel attention $\alpha$.
% Then, each task-specific feature is contextualized as follows:
% \begin{gather}
%     \alpha = \sigma(\text{MLP}([q^{a}_{i};f_{i}]))
% \end{gather}
% The multiple context information is refined using MLP layers conditioned on  each sub-task and applied the channel attention with element-wise multiplication. Formally, the contextualized task-specific feature is written as:
% Then, the relation embedding feature is fused with the task-specific feature as:
% \sh{Formally, the channel attention $\alpha$ is formulated as Eq.~\ref{eq:alpha}.
% \sh{The multiple relation context information is fed into MLP with the task-specific feature to propagate the requisite context information for each sub-task.}
Formally, the channel attention $\alpha$ and the refined tokens $\mathbf{Q}^\tau_{(l)}$ are formulated as follows:
\begin{align}
    \alpha &= \sigma(\text{MLP}([\mathbf{f}^{\tau}_{(l),i};\mathbf{m}_{(l),i}])) \label{eq:alpha}\\
    \mathbf{q}^{\tau}_{(l),i} &= \mathbf{f}^{\tau}_{(l),i} + \alpha \odot \text{MLP}([\mathbf{f}^{\tau}_{(l),i};\mathbf{m}_{(l),i}]),
\end{align}
where we denote $\odot$ and $\sigma$ as element-wise multiplication, and sigmoid function, respectively.
% The contextualized task-specific features are the output of decoders in \ours~(\textit{i.e.,} $Q^a_{(l)}=\{\tilde{q}^{a}_{l,i}\}^{N_q}_{i=1}$).
As the refined tokens $\mathbf{Q}^\tau_{(l)}$ is generated via context exchange with the multiplex relation context, it deduces the comprehensive relational understanding to discover HOI instances.
%the model can predict the HOI triplet via the relation reasoning.}

The $\mathbf{Q}^{\tau}_{(L)}$, the output of last layer of branch $\tau$, is fed into FFNs to predict a set of the HOI predictions.
Formally, given the $\mathbf{Q}^{\tau}_{(L)}$, the MUREN~predicts a set of HOI predictions \{$(\mathbf{b}^{\mathrm{H}}_{i}, \mathbf{b}^{\mathrm{O}}_{i}, \mathbf{p}^{\mathrm{O}}_{i}, \mathbf{p}^{\mathrm{I}}_{i})\}_{i=1}^{N}$ using FFNs as follows:
\begin{align}
    \mathbf{b}^\mathrm{H}_i &= \text{FFN}_{\mathrm{hbox}}(\mathbf{q}^\mathrm{H}_{(L),i}) \in \mathbb{R}^{4}, \\
    \mathbf{b}^\mathrm{O}_i &= \text{FFN}_{\mathrm{obox}}(\mathbf{q}^\mathrm{O}_{(L),i}) \in \mathbb{R}^{4}, \\
    \mathbf{p}^\mathrm{O}_i &= \delta(\text{FFN}_{\mathrm{oc}}(\mathbf{q}^\mathrm{O}_{(L),i})) \in \mathbb{R}^{|\mathcal{O}|},\\
    \mathbf{p}^\mathrm{I}_i &= \sigma(\text{FFN}_{\mathrm{ic}}(\mathbf{q}^\mathrm{I}_{(L),i})) \in \mathbb{R}^{|\mathcal{I}|},
\end{align}
where $\delta$ is a softmax operation, and $\mathbf{p}^\mathrm{O}_i$, $\mathbf{p}^\mathrm{I}_i$ are class probability of object and interaction, respectively. 
% \subsection{Inference}
% Given the outputs of the last decoding branch layer, the \ours~predicts the HOI triplets $\{(b^H_i,\xi^{O}_{i},c^I_i)\}^{N_q}$ using FFN as follows:
% \begin{gather}
%     b^H_i = \text{FFN}_{hbox}(q^H_{i,N_{dec}}), \\
%     b^O_i = \text{FFN}_{obox}(q^O_{i,N_{dec}}), \\
%     c^O_i = \text{FFN}_{oc}(q^O_{i,N_{dec}}), \\
%     c^I_i = \text{FFN}_{ic}(q^I_{i,N_{dec}}),
% \end{gather}
% where $\xi^{O}_{i}=(b^O_i,c^O_i)$, $b^H\in\mathbb{R}^{4}$, $b^O\in\mathbb{R}^{4}$, $c^{O}\in\mathbb{R}^{\mathcal{C}}$ and $c^I\in\mathbb{R}^{\mathcal{I}}$. $\mathcal{C}$, $\mathcal{I}$ indicate the number of object classes and interaction classes, respectively.
% The update query set $Q^{a}_{(l)}$ is utilized as input of $l+1$ decoding branch layer.
% The query set $Q^{a}_{N_{dec}}$, output of last decoding brnach layer, is fed into to predict a set of the HOI triplets  $\{b^{H}_i,\xi^{O}_i,c^I_i\}$   
% 
\subsection{Training Objective}
% each feature predicts the task in charge of compose the HOI triplet.
% Specifically, the $\tilde{q}^{H}_{i}$ is responsible for detecting the human box $b_i^H\in\mathbb{R}^{4}$.
% The $\tilde{q}^{O}_{i}$ is responsible for predicting the object box $b_i^O\in\mathbb{R}^{4}$ and the object class $c^O\in\mathbb{R}^{\mathcal{C}}$, where $\mathcal{C}$ is the number of the object classes.
% The $\tilde{q}^{I}_{i}$ is responsible for predicting the interaction class $c^{I}_i\in\mathbb{R}^{\mathcal{I}}$, where $\mathcal{I}$ is the number of the interaction classes.
% We group the predictions of the same index $i$ together to compose an HOI triplet prediction $\{b^{H}_i,b^O_{i},c^{O}_i,c^I_i\}$.
% the \ours~predicts a set of HOI triplet $\{b^{H}_i,b^O_{i},c^{O}_i,c^I_i\}$ for all query index $i$:
% % where $b^H_i = \text{FFN}_{\text{hbox}}(\tilde{q}^H_i)$, $b^O_i = \text{FFN}_{\text{obox}}(\tilde{q}^O_i$), $c^{O}_i = \text{FFN}_{\text{ocls}}(\tilde{q}^O_i)$ and $c^{I}_i = \text{FFN}_{\text{icls}}(\tilde{q}^I_i)$.
% % \begi}
% %     b^H_i = \text{FFN}_{\text{hbox}}(\tilde{q}^H_i), \\
% %     b^O_i = \text{FFN}_{\text{obox}}(\tilde{q}^O_i),  \\
% %     c^{O}_i = \text{FFN}_{\text{ocls}}(\tilde{q}^O_i),  \\
% %     c^{I}_i = \text{FFN}_{\text{icls}}(\tilde{q}^I_i),
% \end{gather}
% where $b^H\in\mathbb{R}^{4}$, $b^O\in\mathbb{R}^{4}$, $c^{O}\in\mathbb{R}^{\mathcal{C}}$, $c^I\in\mathbb{R}^{\mathcal{I}}$, and $\mathcal{C}$, $\mathcal{I}$ indicate the number of object class and the number of interaction class, respectively.
For training our proposed method, we follow previous transformer-based methods~\cite{tamura2021qpic,zhang2021cdn,zhou2022distr}. 
We adopt the Hungarian Matching~\cite{kuhn1955hungarian} to assign the ground-truth HOI instances to the predictions.
MUREN is trained with multi-task loss composed of four losses: L1 loss~\cite{ren2015faster} $\mathcal{L}_\mathrm{L1}$ and GIoU loss~\cite{rezatofighi2019generalized} $\mathcal{L}_\mathrm{GIoU}$ for the bounding box regression, cross-entropy loss $\mathcal{L}_{oc}$ for the object classification, and focal loss~\cite{lin2017focal} $\mathcal{L}_{ic}$ for the interaction classification. 
% \ours~is trained with multi-task losses, which contain the bounding box L1 loss $\mathcal{L}_{box}$, the GIoU loss $\mathcal{L}_{giou}$, the object classification loss $\mathcal{L}_{oc}$, and the interaction classification loss $\mathcal{L}_{ic}$.
% We use the cross-entropy for computing $\mathcal{L}_{oc}$ and the focal loss~\cite{lin2017focal} for $\mathcal{L}_{ic}$.
% Following the previous work~\cite{tamura2021qpic}, \ours~is trained with multi-task losses, which contain the bounding box L1 loss $\mathcal{L}_{box}$, the GIoU loss $\mathcal{L}_{giou}$, the object classification $\mathcal{L}_{oc}$ and the interaction classification loss $\mathcal{L}_{ic}$. The Hungarian Matching~\cite{kuhn1955hungarian} is used to assign the ground-truth HOI triplets to the predictions.
The total loss $\mathcal{L}$ is formulated as:
\begin{equation}
    \mathcal{L} = \lambda_{\mathrm{L1}}\mathcal{L}_{\mathrm{L1}} + \lambda_{\mathrm{GIoU}}\mathcal{L}_{\mathrm{GIoU}} + \lambda_{\mathrm{oc}}\mathcal{L}_{\mathrm{oc}} + \lambda_{\mathrm{ic}}\mathcal{L}_{\mathrm{ic}}, 
\end{equation}
% \begin{gather}
%     \mathcal{L} = \sum\limits_{a}(\lambda^{a}_{box}\mathcal{L}^{a}_{box} + \lambda^{a}_{giou}\mathcal{L}^{a}_{giou}) + \lambda_{oc}\mathcal{L}_{oc} + \lambda_{ic}\mathcal{L}_{ic}, 
% \end{gather} 
where $\lambda_\mathrm{L1}$, $\lambda_\mathrm{GIoU}$, $\lambda_{\mathrm{oc}}$, and $\lambda_{\mathrm{ic}}$ are the hyper-parameters for weighting each loss.
% \sh{The matching cost for the bipartite matching process is same as the total loss $\mathcal{L}$.} 
Additionally, we apply intermediate supervision for better representation learning.
Specifically, we attach the same FFNs to each decoding branch layer to calculate the intermediate loss.
This auxiliary loss is computed the same as $\mathcal{L}$.
% For more details about the Hunagarian Matching and each loss, please refer to~\cite{tamura2021qpic}. 

% equip each decoder layer with several FFN heads for intermediate
% supervision. 

\subsection{Inference}
 \sh{Given the set of HOI predictions, we generate a set of HOI instances $\{(\mathbf{b}^{\mathrm{H}}_{i}, \mathbf{b}^{\mathrm{O}}_{i}, \mathbf{c}^{\mathrm{O}}_{i,j'}, \mathbf{c}^{\mathrm{I}}_{i,t}) |~i\in N,~k \in \mathbb{R}^{|\mathcal{I}|},~j'=\mathrm{argmax}_{j}\mathbf{p}^{\mathrm{O}}_{i,j} \}$, where $c^{O}_{i,j'}  \in \mathbb{R}^{|\mathcal{O}|}$, $c^{I}_{i,t}  \in \mathbb{R}^{|\mathcal{I}|}$ are one-hot vectors with the $j$-th and $t$-th index set to 1, respectively. Following ~\cite{zhang2021cdn}, we then select top-$k$ score HOI instances, where the score is given by $ \mathbf{p}^{\mathrm{O}}_{i,j'} \cdot \mathbf{p}^{\mathrm{I}}_{i,t}$.}

% !TEX root = ../main.tex
\section{Experiments}
\subsection{Datasets and Metrics}
We evaluate our model on the two public benchmark datasets: HICO-DET~\cite{hico} and V-COCO~\cite{vcoco}. 

\noindent
\textbf{HICO-DET} has 38,118 images for training and 9,658 images for testing.
It contains 80 object classes, 117 interaction classes and 600 HOI classes, which are a pair of an object class and an interaction class (\textit{e.g.,} `riding bicycle'). 
We evaluate the proposed method on Default and Known Object settings.
% The mAP is calculated over all testing images in Default setting.
In the Default setting, the AP is calculated across all testing images for each HOI class.
The Known Object setting calculates the AP of an HOI class over the images containing the object in the HOI class (\textit{e.g.,} the AP of an HOI class `riding bicycle' is only calculated on the images which contain the object `bicycle').
% measures the AP of each object solely over the images containing that object class.
% Default
% setting: For each HOI category, we evaluate the detection
% on the full test set, including images both containing and not
% containing the target object category. 
% Known
% Object setting: For each HOI category (e.g. “riding a
% bike”), we evaluate the detection only on the images containing the target object category (e.g. “bike”). The challenge is to localize HOI (e.g. human-bike pairs) as well as
% distinguishing the interaction (e.g. “riding”).
% Default
% setting represents that the mAP is calculated over all testing images, while Known Object measures the AP of each
% object solely over the images containing that object class.
Following the previous work~\cite{zhang2021cdn}, we report the mAP under three splits (Full, Rare, and Non-Rare) for each setting.
The Full, Rare, and Non-Rare splits contain all 600 HOI classes, 138 HOI classes, which have less than 10 training samples for each class, and 462 HOI classes, which have more than 10 training samples for each class, respectively.

\noindent
\textbf{V-COCO} is a subset of the MS-COCO~\cite{coco} dataset. It consists of 5400 and 4,946 images for training, and testing.
It has 80 object classes and 29 action classes. 
Following the evaluation settings in ~\cite{kim2021hotr}, we evaluate the proposed method on scenario 1 and scenario 2, and report role average precision under two scenarios ($\mathrm{AP}^{\#1}_{\mathrm{role}}$ for scenario 1 and $\mathrm{AP}^{\#2}_{\mathrm{role}}$ for scenario 2).
In scenario 1, the model should predict the bounding box of the occluded object as [0,0,0,0].
In contrast, the predicted bounding box of the occluded object is ignored on calculating the $\mathrm{AP}_{\mathrm{role}}$ in scenario 2.
%  Specifically, in the scenario of AP #1
% role,
% the model should manage to infer the occluded object correctly by predicting the 2D location of its bounding box as
% [0,0,0,0], meanwhile precisely localizing the corresponding
% human bounding box and recognizing the interaction in between. In contrast, for the scenario of AP #2
% role, there is no
% need to infer the occluded object.

\vspace{-1mm}
\subsection{Implementation Details}
The encoder in MUREN~adopts ResNet-50 as a CNN backbone followed by a 6-layer transformer encoder.
We set the number of branch layers $L$ to 6.
For the training, we set the number of queries $N$ to 64 for HICO-DET and 100 for V-COCO following~\cite{zhang2021cdn}.
The weight of loss $\lambda_{\mathrm{L1}}$, $\lambda_{\mathrm{GIoU}}$, $\lambda_{oc}$, $\lambda_{ic}$ is set to 2.5, 1, 1, 1, respectively.
The network is initialized with the parameters of DETR~\cite{detr} pretrained on MS-COCO~\cite{coco}.
We optimize our network by AdamW~\cite{loshchilov2017decoupled} with the weight decay $1e{-4}$. 
We set the initial learning rate of the CNN backbone to $1e{-5}$ and the other component to $1e{-4}$. The model is trained with 100 epoch.
For the V-COCO, we freeze the CNN backbone to prevent overfitting, and set the learning rate to $4e{-5}$. All experiments are conducted with a batch size of 16 on 4 RTX 3090 GPUs.

% The encoder of \ours~adopts Resnet-50 CNN backbone with 6-layer transformer encoder. We set the number of query $N_q$ to 64 for HICO-DET and 100 for V-COCO, and the number of decoder layer $N_{dec}$ to 6. The weight of loss $\lambda_{box}$, $\lambda_{giou}$, $\lambda_{oc}$, $\lambda_{ic}$ is set to 2.5, 1, 1, 1, respectively. fThe network is initialized with the parameters of DETR~\cite{detr} pretrained on MSCOCO dataset. We optimize the network by AdamW with the batch size 16, the learning rate $1e^{-4}$ for the encoder and decoder, $1e^{-5}$ for the CNN backbone. For the V-COCO, we freeze the CNN backbone, set the learning rate $4e^{-5}$ for the encoder and decoder, and share the weight of attentive fusion module at each decoder branch to prevent over-fitting.

\begin{table*}[!t]
\begin{center}
\scalebox{0.82}{
\begin{tabular}{ccc P{1.4cm}P{1.4cm}P{1.4cm} c P{1.4cm}P{1.4cm}P{1.4cm}}
\toprule 
\multirow{2}{*}{Method} & \multirow{2}{*}{Backbone} & \multirow{2}{*}{Feature} & \multicolumn{3}{c}{Default} && \multicolumn{3}{c}{Known Object} \\
&&& Full & Rare & Non-Rare && Full & Rare & Non-Rare \\
\midrule
\multicolumn{10}{l}{\textbf{CNN-based methods}} \\
\midrule
iCAN~\cite{gao2018ican} & R50 & A+S & 14.84 & 10.45 & 16.15 && 16.26 & 11.33 & 17.73  \\
TIN~\cite{li2019tin} & R50 & A+S+P & 22.90 & 14.97 & 25.26 && - & - & - \\
GPNN~\cite{qi2018dpnn} & R101 & A & 13.11 & 9.34 & 14.23 && - & -  & -\\
DRG~\cite{gao2020drg} & R50-FPN & A+S+L+M & 24.53 & 19.47 & 26.04 && 27.98 & 23.11 & 29.43 \\
VSGNet~\cite{ulutan2020vsgnet} & R152 & A+S & 19.80 & 16.05 & 20.91 && - & - & - \\
wang \textit{et al}.~\cite{wang2020hetero} & R50-FPN & A+S+M & 17.57 &16.85& 17.78&&21.00& 20.74 &21.08 \\
IDN~\cite{li2020IDN} & R50 & A+S & 26.29 & 22.61 & 27.39 && 28.24 & 24.47 & 29.37 \\ 
VCL~\cite{hou2020vcl} & R50 &  A & 23.63 & 17.21 & 25.55 && 25.98 & 19.12 & 28.03 \\
UnionDet~\cite{kim2020uniondet} & R50 & A & 17.58 & 11.72 & 19.33 && 19.76 & 14.68 & 21.27\\
GGNet~\cite{zhong2021glance} & HG104 & A & 28.83 & 22.13 & 30.84 && 27.36 & 20.23 &  29.48 \\
SCG~\cite{zhang2021spatially} & R50-FPN & A+S+M & 31.33 & 24.72 & 33.31 && 34.37 & 27.18 & 36.52 \\

\midrule
\multicolumn{5}{l}{\textbf{Transformer-based methods}} \\
\midrule
PST~\cite{dong2021pst} & R50 & A & 23.93 & 14.98 & 26.60 && 26.42 & 17.61 & 29.05 \\
HoiTrans~\cite{zou2021hoitrans} & R101 & A & 26.61 & 19.15 & 28.84 && 29.13 & 20.98 & 31.57 \\ 
HOTR~\cite{kim2021hotr} & R50 & A  & 25.10 & 17.34 & 27.42 && - & - & - \\
AS-Net~\cite{chen2021asnet} & R50 & A & 28.87 & 24.25 & 30.25 && 31.74 & 27.07 & 33.14 \\
QPIC~\cite{tamura2021qpic} & R101 & A & 29.90& 23.92& 31.69&& 32.38& 26.06 & 34.27 \\
MSTR~\cite{kim2022mstr} & R50 & A+M & 31.17 & 25.31 & 32.92 && 34.02 & 28.83 & 35.57 \\
CDN~\cite{zhang2021cdn} & R101 & A & 32.07 & 27.19 & \underline{33.53} && 34.79 & 29.48 & 36.38 \\
UPT~\cite{zhang2022upt} & R50 & A+S & 31.66 & 25.94 & 33.36 && 35.05 & 29.27 & \underline{36.77} \\
DisTR~\cite{zhou2022distr} & R50 & A & 31.75 & 27.45 & 33.03 && 34.50 & 30.13 & 35.81 \\
%STIP~\cite{zhang2022stip} & R50 & A & 31.60 & 27.75 & 32.75 && 34.41 & 30.12 & 35.69 \\
STIP~\cite{zhang2022stip} & R50 & A+S+L & \underline{32.22} & \underline{28.15} & 33.43 && \underline{35.29} & \textbf{31.43} & 36.45 \\  
\hline \midrule
Ours & R50 & A & \textbf{32.87} & \textbf{28.67} & \textbf{34.12} && \textbf{35.52} & \underline{30.88} & \textbf{36.91} \\
\bottomrule
\end{tabular}}
\end{center}
\vspace{-0.45cm}
\caption{Performance comparison on the HICO-DET~\cite{hico} dataset. The letters in Feature column stand for A: Appearance/Visual features, S: Spatial features, L: Linguistic features, P: Human pose features, M: Multi-scale features.
The best score is highlighted in bold, and the second-best score is underscored.}
\vspace{-0.4cm}
\label{tab:hicodet-res}
\end{table*}

\vspace{-1mm}
\subsection{Comparison with State-of-the-Art}
Table~\ref{tab:hicodet-res} and Table~\ref{tab:vcoco-res} show the performance comparison of the proposed method with the previous HOI methods.
As shown in Table~\ref{tab:hicodet-res}, on the HICO-DET dataset, the proposed method achieves state-of-the-art performance on Default and Known Object settings against existing CNN- and transformer-based methods.
% In particular, we outperform the previous entangled transformer-based methods~\cite{zou2021hoitrans,tamura2021qpic,kim2022mstr} by a large margin.
% Furthermore, \ours~shows better performance than the previous disentangled transformer-based methods~\cite{zhang2022upt,zhang2021cdn,dong2021pst,zhou2022distr,chen2021asnet}.
% the conventional two-stage HOI detectors
% (e.g., GPNN, TIN, DRG) commonly construct instancecentric graph to mine contextual information among instances
Compared with the previous CNN-based methods~\cite{zhang2021spatially,qi2018dpnn,gao2020drg,ulutan2020vsgnet,wang2020hetero}, which utilize the graph structure for context exchange, MUREN~shows significant improvements.
We also surpass the previous single-branch methods~\cite{zou2021hoitrans,kim2022mstr,tamura2021qpic}.
It illustrates that it is crucial extracting the task-specific tokens for each sub-task with different branches.
In particular, we outperform the previous two-branch methods~\cite{zhang2022upt,zhang2021cdn,zhou2022distr,chen2021asnet,kim2021hotr}.
DisTR~\cite{zhou2022distr} and AS-NET~\cite{chen2021asnet} perform context exchange for relational reasoning, but they only propagate the context information of the human and the object to the interaction branch for interaction classification.
% Although PST~\cite{dong2021pst} propagates the context information to each sub-task, the context information is limited to the unary or the triplet relation context information.
% Although PST~\cite{dong2021pst} propagates the context information to the features for each sub-task, the context information is limited to one type of relation context information.
Instead, we exchange the context information among the three branches, selecting requisite context information from the multiplex relation context for each sub-task.
% , which contains the unary, pairwise, and triplet relation context information, and perform the context exchange among three branches with the multiple relation context information for relational reasoning.
% and tree-branch architecture to extract the discriminative features for each sub-task
These results illustrate the advantage of context exchange between each branch using the multiplex relation context for relational reasoning.
Moreover, MUREN~shows better performance without using any additional information (\textit{e.g.,} spatial and linguistic information) compared with~\cite{kim2022mstr,zhang2022stip,zhang2022upt,zhang2021spatially}.
We also outperform~\cite{zhang2021cdn,tamura2021qpic,zou2021hoitrans} which utilize a deeper backbone to extract discriminative features for each sub-task.
These results illustrate that three-branch architecture and context exchange with multiplex relation context for relational reasoning provide more discriminative features to predict each sub-task.
% (\textit{e.g.,} spatial and linguistic features),~\cite{kim2022mstr,zhang2022stip,zhang2022upt,zhang2021spatially}.
We further evaluate MUREN~on the V-COCO dataset and observe similar results as in the HICO-DET dataset.
As shown in Table~\ref{tab:vcoco-res}, MUREN~achieves state-of-the-art performances across all the metrics compared with existing methods.
% The results illustrate the effectiveness of the three-branch structure and multiple relation context information for the conrtext exchange. 
% Similar to the observations on V-COCO,
% our STIP achieves consistent performance gains against existing HOI detectors across all the metrics for each training
% setting. The results basically demonstrate the advantage of
% triggering HOI set prediction with the non-parametric interaction proposals and meanwhile exploiting the holistically
% semantic structure among interaction proposals & the locally spatial structure within each interaction proposal.

% Compared with previous two-stage methods~\cite{qi2018dpnn,ulutan2020vsgnet,zhang2021spatially,wang2020hetero,gao2020drg}, which use the message passing mechanism, we outperforms  

% Compared with previous the entangeled-
% We also outperforms the previous transformer-based methods.
% Especially, compared with previous disentangled transformer-based methods methods~\cite{chen2021asnet,kim2021hotr,zhang2021cdn,dong2021pst,zhou2022distr}.

\begin{table}[t]
\begin{center}
\scalebox{0.80}{
\begin{tabular}{ccccc}
\toprule
Method & Backbone & Feature & $\mathrm{AP}^{\#1}_{\mathrm{role}}$ & $\mathrm{AP}^{\#2}_{\mathrm{role}}$ \\
\midrule
\multicolumn{5}{l}{\textbf{CNN-based methods}} \\
\midrule
GPNN~\cite{qi2018dpnn}& R101 & A &  44.0 & - \\
iCAN~\cite{gao2018ican}& R50 &A+S& 45.3 & 52.4 \\
TIN~\cite{li2019tin} & R50 & A+S+P & 47.8 & 54.2 \\
VSGNet~\cite{ulutan2020vsgnet} & R152 & A+S & 51.8 & 57.0 \\
DRG~\cite{gao2020drg} & R50-FPN & A+S+L+M & 51.0 & - \\
VCL~\cite{hou2020vcl} & R101 & A & 48.3 & - \\ 
UnionDet~\cite{kim2020uniondet} & R50 & A & 47.5 & 56.2\\
GGNet~\cite{zhong2021glance}& HG104 & A & 54.7& -  \\
IDN~\cite{li2020IDN} & R50 & A+S & 53.3 & 60.3 \\ 
SCG~\cite{zhang2021spatially} & R50-FPN & A+S+M & 54.2 & 60.9 \\
\midrule
\multicolumn{5}{l}{\textbf{Transformer-based methods}} \\
\midrule
% Transformer-based methods \\
QPIC~\cite{tamura2021qpic} & R50 & A & 58.8 & 61.0 \\
MSTR~\cite{kim2022mstr} & R50 & A+M & 62.0 & 65.2 \\
HOTR~\cite{kim2021hotr} & R50 & A & 55.2 & 61.0 \\
AS-NET~\cite{chen2021asnet} & R50 & A & 53.9 & - \\
CDN~\cite{zhang2021cdn} & R101 & A & 63.9 & 65.9\\
UPT~\cite{zhang2022upt} & R50 & A & 59.0 & 64.5  \\
STIP~\cite{zhang2022stip} & R50 & A+S+L &  66.0 & \underline{70.7} \\  
DisTR~\cite{zhou2022distr} & R50 & A & \underline{66.2} & 68.5  \\
\hline\midrule
Ours & R50 & A &  \textbf{68.8} & \textbf{71.0} \\
% Ours & R101 & 69.64 & 72.44 \\
\bottomrule
\end{tabular}
}
\end{center}
\vspace{-0.4cm}
\caption{Performance comparison on V-COCO~\cite{vcoco} dataset. The letters in Feature column stand for A: Appearance/Visual features, S: Spatial features, L: Linguistic features, P: Human pose features, M: Multi-scale features.
The best score is highlighted in bold, and the second-best score is underscored.}
\vspace{-0.6cm}
\label{tab:vcoco-res}
\end{table}

\vspace{-1.5mm}
\subsection{Ablation Study}
We conduct various ablation studies on the V-COCO dataset to validate the effectiveness of~MUREN.

\noindent
\textbf{Impact of each relation context information on relational reasoning.}
We utilize the multiplex relation context, which contains the unary, pairwise, and ternary relation context, for relational reasoning.
To investigate the impact of each relation context information on relational reasoning, 
we gradually add each relation context information to the baseline, which predicts the HOI instances without context exchange among each branch for relational reasoning.
As shown in Table~\ref{tab:ablation-multiplex-res}, we observe that context exchange using the ternary relation context gives 4.55\%p, 4.22\%p improvement with a large margin in $\mathrm{AP}^{\#1}_{\mathrm{role}}$ and $\mathrm{AP}^{\#2}_{\mathrm{role}}$, respectively.
% \sh{This result indicates that context exchange with ternary relation context information among the three branches promotes relational reasoning to discover the HOI instances since the ternary relation contains the holistic context information about the HOI instances.}
This result indicates that context exchange for relational reasoning is essential for discovering the HOI instance and ternary relation context promotes relational reasoning providing holistic information about the HOI instances.
Besides, when the model exploits ternary and unary relation contexts, the model shows an additional performance improvement.
We observe similar results on the model which utilizes both ternary and pairwise relation contexts.
% It indicates that the fine-grained relation context information enriches the triplet relation information and helps the relation reasoning to predict the HOI triplet.
It indicates that the fine-grained relation contexts provide useful information for relational reasoning to predict HOI instances.
When we use all the relation context information in HOI instance, the model shows a significant performance increase of 6.23\%p and 5.86\%p in $\mathrm{AP}^{\#1}_{\mathrm{role}}$ and $\mathrm{AP}^{\#2}_{\mathrm{role}}$, compared with the baseline.
It demonstrates that each relation context information complements the others, and thus the multiplex relation context provides rich information for relational reasoning and brings performance gain in HOI detection.

\noindent
\textbf{Impact of the multiplex relation context on each sub-task.}
% We investigate the impact of the context exchange among the sub-tasks.
For investigating the propagation impact of the multiplex relation context on the sub-tasks, we gradually add the propagation the multiplex relation context to each branch.
When we propagate the multiplex relation context to one of the detection branches (\textit{i.e.,} human branch and object branch), we observe that the model consistently shows performance improvement compared with the baseline, as shown in Table~\ref{tab:ablation-message-res}.
We also observe the performance gains when the model propagates the multiplex relation context to both human and object branch.
It indicates that relational context information is required to detect the human and the object in the HOI detection.
In particular, when the model propagates the multiplex relation context to the interaction branch, MUREN~shows the notable performance gains of 3.19\%p and 2.77\%p on scenario 1 and scenario 2.
It indicates that the multiplex relation context is essential to interaction classification which requires a comprehensive relational understanding between the human and the object.
% we observe the notable performance gain %since it requires comprehensive visual understanding to infer the interaction between the human and the object.
The entire model of MUREN, which propagates the relation context information to all sub-tasks, achieves the highest performance with a significant margin compared with the other model variants.
The results demonstrate that context exchange among the three branches is essential to identify HOI instances and plays a crucial role in the comprehensive relational understanding.
%the propagation of the multiple relation context information on each sub-tasks
% It demonstrates that the multiple relation context information plays a crucial role in predicting each sub-task and the multiplex relation context information contains the context information about each-sub task, boosting the performance of each sub-task.

\begin{table}[]
\begin{center}
\scalebox{0.88}{
\begin{tabular}{ccccccc}
\toprule
ternary & unary & pairwise & $\mathrm{AP}^{\#1}_{\mathrm{role}}$ & $\mathrm{AP}^{\#2}_{\mathrm{role}}$ \\
\midrule
- & - & - &  62.52 & 65.14  \\
\checkmark & - & -  &  67.07 & 69.36  \\
\checkmark & \checkmark & - &  68.12 & 70.31  \\
\checkmark & - &\checkmark & 67.67 & 70.02   \\
\hline\midrule
\checkmark & \checkmark &\checkmark & \textbf{68.75} & \textbf{71.00}  \\
\bottomrule
\end{tabular}}
\end{center}
\vspace{-0.4cm}
\caption{The impact of each relation context information on relational reasoning. The `ternary', `unary', and `pairwise' columns indicate the ternary, unary and pairwise relation context.}
\label{tab:ablation-multiplex-res}
\end{table}

\begin{table}[]
\begin{center}
\scalebox{0.82}{
\begin{tabular}{cccccc}
\toprule
human & object & interaction & $\mathrm{AP}^{\#1}_{\mathrm{role}}$ & $\mathrm{AP}^{\#2}_{\mathrm{role}}$ \\
\midrule
- & - & - &  62.52 & 65.14 \\
\checkmark & - & - &  64.44	& 66.62 \\
- & \checkmark & - &   63.66 & 66.00  \\
\checkmark & \checkmark & - &  65.29 & 67.5 \\
- & - &\checkmark &  65.71 & 67.91  \\
%\checkmark & - &\checkmark &  66.35	& 68.57  \\
%- & \checkmark &\checkmark & 66.55 & 68.81\\
\hline\midrule
\checkmark & \checkmark &\checkmark & \textbf{68.75} & \textbf{71.00} \\
\bottomrule
\end{tabular}}
\end{center}
\vspace{-0.4cm}
\caption{The impact of the multiplex relation context on each sub-task. The `human', `object', and `interaction' columns indicate the propagation of the multiplex relation context to human, object, and interaction branch, respectively.}
\vspace{-2mm}
\label{tab:ablation-message-res}
\end{table}

\begin{table}[]
\begin{center}
\scalebox{0.82}{
\begin{tabular}{cccc}
\toprule
conditioning & channel  & $\mathrm{AP}^{\#1}_{\mathrm{role}}$ & $\mathrm{AP}^{\#2}_{\mathrm{role}}$ \\
\midrule
- & - & 66.50 & 68.96 \\
\checkmark & - & 66.95 & 69.23 \\
- & \checkmark & 67.10  &  69.49 \\
\hline\midrule
\checkmark & \checkmark & \textbf{68.75} &  \textbf{71.00} \\
\bottomrule
\end{tabular}}
\end{center}
\vspace{-0.4cm}
\caption{Ablations studies on each component in the attentive fusion module. `conditioning` and `channel` indicate transforming multiplex relation context conditioned on a task-specific token} and channel attention mechanism.
\label{tab:ablation-att-res}
\vspace{-0.6cm}
\end{table}

\noindent
\textbf{Impact of attentive fusion module on context exchange.}
MUREN~exchanges relational context information between each branch via the attentive fusion module.
% the multiplex relation context information to each task-specific feature via the attentive fusion module. 
To investigate the impact of the attentive fusion module, we remove the attentive fusion module and fuse both the task-specific tokens and the multiplex relation context with an element-wise addition operation for the baseline.
As shown in Table~\ref{tab:ablation-att-res}, the performance drops by 2.25\%p and 2.04\%p in the two scenarios.
It shows the effectiveness of our attentive fusion module for context exchange between the branches.

% \vspace{1mm}
\noindent
\textbf{Impact of the context information selection for each sub-task.}
In the attentive fusion module, we select requisite context information for each sub-task from the multiplex relation context.
% In the attentive fusion module, we select the context information to propagate the requisite context information for each sub-task.
We further analyze the impact of the context information selection as shown in Table~\ref{tab:ablation-att-res}.
To select the requisite context information for each sub-task, we utilize 1) transforming
multiplex relation context conditioned on a task-specific token (`conditioning' in Table~\ref{tab:ablation-att-res}) and 2) channel attention mechanism (`channel' in Table~\ref{tab:ablation-att-res}).
We observe that the model, which utilizes one of `conditioning' and `channel', gains performance improvement.
We also observe that the model with both `conditioning' and `channel' shows better performance than the other model variants.
The results demonstrate that each sub-task requires different context information for relational reasoning, and thus it is important to propagate the requisite context for each sub-task.
Our attentive fusion module effectively selects requisite context information for each sub-task.

\noindent
\textbf{Impact of disentangling human and object branches.}
Human plays a central and an active role for HOI, which is distinctive from a relatively passive role of object, and thus requires a dedicated module to capture relevant attributes and semantics such as pose and clothing. 
%Thus, capturing these subtle semantics of the people is crucial for HOI detection.
% To demonstrate the advantages of decoupling the human and object features,
We evaluated in Table~\ref{tab:ablation-two-branch}
the effect of sharing parameters between human and object branches; we gradually increased the number of layers that share parameters between the two branches.
The results show that increasing the number of shared layers drops the performance and the full-sharing model, MUREN-(6), results in 2.2\%p and 1.9\%p decrease in performance at two scenarios, respectively, compared with non-sharing model, MUREN-(0).
%separation was increased (i.e. sharing the parameters only in front of three layers), MUREN-(3) exhibited a 0.89 and 0.79 increase in performance in the two scenarios compared to MUREN-(6).
This is a significant drop also compared to MUREN$^{\dagger}$, which has a similar number of parameters with MUREN-(6) by adjusting the number of layer $L$ of MUREN, indicating that separating human and object branches is important indeed for HOI detection.

\begin{table}[!]
\begin{center}
\scalebox{0.82}{
\begin{tabular}{cccc}
\toprule
Method & $\mathrm{AP}^{\#1}_{\mathrm{role}}$ & $\mathrm{AP}^{\#2}_{\mathrm{role}}$ & Params (M) \\
\midrule
MUREN-(0) & 68.8 & 71.0 & 69.3 \\
%MUREN$^{\ddagger}$-(0) & 68.27 & 70.62 & 59.59 \\
MUREN-(3) & 67.1 & 69.3 & 64.3 \\
MUREN-(6) & 66.6 & 69.1 & 59.6 \\
MUREN$^{\dagger}$ & 68.3 & 70.6 & 59.6 \\ 
% -intermediate loss& & 64.74 & 67.10 \\
% \midrule
% BASE & 62.52 & 65.14\\
% + tenary & 67.07 & 69.36\\
% + unary & 67.01 & 6.9.31 \\
% + pairwise &  67.70(67.41) & 70.13 \\
% + tenary^{\dagger} & 67.22(67.36) & 69.51(69.68) \\
\bottomrule
\end{tabular}
}

\end{center}
\vspace{-0.4cm}
\caption{The Impact of disentangling human and object branches. MUREN-($k$) denotes the sharing of parameters between the human and object branches across $k$ layers. The parameters are shared only between corresponding layers.  MUREN$^{\dagger}$ is variant of MUREN by adjusting the number of layer $L$.} 
\label{tab:ablation-two-branch}
\vspace{-0.5cm}
\end{table}

\subsection{Qualitative Results}
\begin{figure}[!]
\begin{center}
% \hspace*{-0.6cm}
  \includegraphics[width=0.432\textwidth]{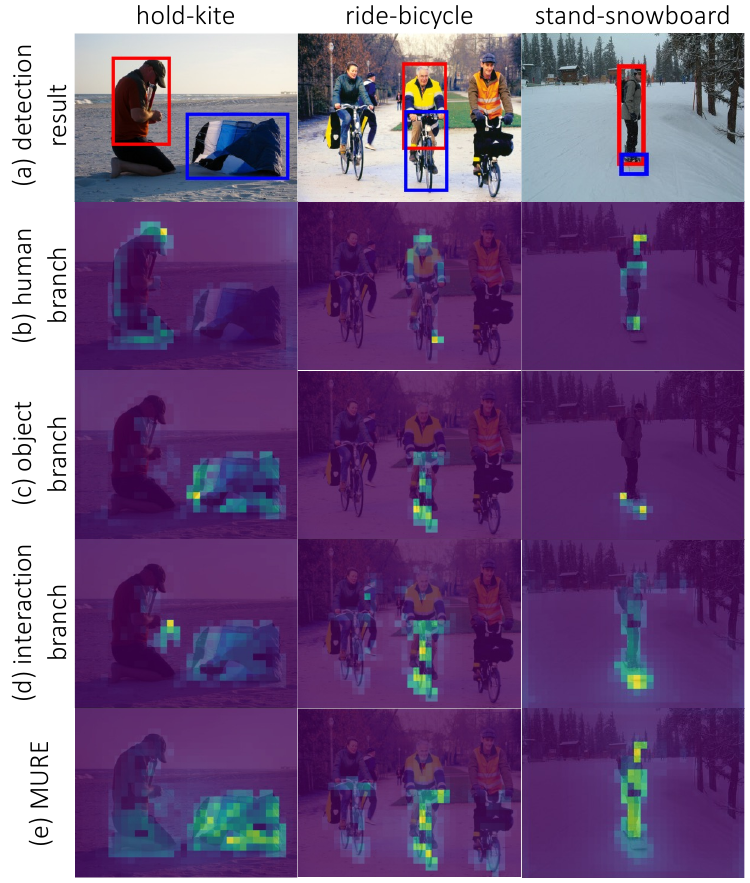}
\end{center}
%   \caption{The visualization on the cross attention map in the each decoder branch and the multiple relation context embedding~(MURE) module. The first column shows the detected HOI triplet. The second~firth column show the cross attention map of the transformer decoder in the last branch layer, generated by Eq.~\ref{eq:task-specific}. The last column is the cross attention map of MURE genreated by Eq.~\ref{eq:mure_image_feat}. Best viewed in color.}
\vspace{-0.45cm}
\caption{The visualization of the HOI detection results and the cross-attention map in each branch and the multiplex relation embedding module~(MURE). Best viewed in color.}

% (a) the detected HOI instance. (b)-(d): the cross-attention map of the transformer decoder in the last branch layer, generated by Eq.~\ref{eq:task-specific}. (e) the cross-attention map of MURE in the last branch layer, generated by Eq.~\ref{eq:mure_image_feat}.
% the images in the top two rows are from HICO-DET~\cite{hico} and an image at the bottom  row is from V-COCO~\cite{vcoco}.
\vspace{-0.5cm}
\label{fig:vis_att}
\end{figure}

We visualize HOI detection results and the cross attention map of each branch and the multiplex relation embedding module~(MURE) in Fig.~\ref{fig:vis_att}.
As shown in Fig.~\ref{fig:vis_att}b, c, the human and the object branches focus on the instance extremities to detect the human and the object.
In the Fig.~\ref{fig:vis_att}d, we observe that the interaction branch attends to the regions where the interaction exists between the human and the object.
These results indicate that the task-specific tokens contain context information for predicting each sub-task.
We also observe that the cross-attention map in MURE highlights the overall region that contains the relational semantics about the HOI instance as shown in Fig.~\ref{fig:vis_att}e.
It demonstrates that MURE captures the context information about HOI instance for relational reasoning.
% Moreover, the proposed method properly detects the HOI instance in complicated scene as shown in the second row of Fig.~\ref{fig:vis_att}.
% These qualitative results demonstrate that our method deduces a comprehensive relational understanding of the HOI instance via context exchange with the multiplex relation context information.

% interpret the context information in the task-specific features and the multiple relation context information. 
% As shown in the second~(`Human') and third column~(`Object') in Figure~\ref{fig:vis_att}, The human and the object decoder branches focus on the instance extremities.
% In the fourth column~(`Interaction') of Figure~\ref{fig:vis_att}, we observe that the transformer decoder branch for interaction classification attends to the regions where the interaction is exist between the human and the object.
% It demonstrate that each target-specific features contains the context information which they are responsible for.
% Moreover, we observe that the attention map in MURE module highlights the overall region that covers the component in the HOI instance as shown in last column in Figure~\ref{fig:vis_att}.
% \sh{Additionaly, our proposed omodel }

% !TEX root = ../main.tex
\section{Conclusion}
% We have proposed \ours~that effectively leverages the multiplex relation context information for relational reasoning.
\sh{We have proposed MUREN, a one-stage method that effectively performs the context exchange between the three branches for HOI detection. 
% The proposed MURE module leverages the relation contexts for relational reasoning, while the attention fusion module selects requisite context information for each sub-task.
% The proposed attention fusion module selects requisite context information for each sub-task and propagates it to each branch for context exchange.
% With the three-branch architecture, \ours~learns the discriminative features to predict each sub-task, thus enabling a comprehensive understanding of HOI instances. 
% The three-branch architecture extracts the discriminative features to predict each sub-task.
By leveraging relation contexts for relational reasoning in MURE and using the attention fusion module to select requisite context information for each sub-task, MUREN can learn discriminative features to predict each sub-task.
% The extensive experiments demonstrate the effectiveness of the MUREN, and its component, showing the importance of context exchange between the branches in the one-stage HOI method. MUREN achieves state-of-the-art performance on both HICO-DET and V-COCO benchmarks.
Our extensive experiments demonstrate the importance of context exchange between the branches and the effectiveness of MUREN, which achieves state-of-the-art performance on both HICO-DET and V-COCO benchmarks and its components.
}
% \vspace{-2mm}

%\small{
% \paragraph{Acknowledgements.} 
\noindent \textbf{Acknowledgements.}
This work was supported by the IITP grants (2021-0-00537: Visual common sense through self-supervised learning for restoration of invisible parts in images (50\%), 2021-0-02068: AI Innovation Hub (40\%), and 2019-0-01906: AI graduate school program at POSTECH (10\%)) funded by the Korea government (MSIT).
{\small
\bibliographystyle{ieee_fullname}
\bibliography{egbib}
}

\end{document}